\documentclass{article}

\usepackage[final]{neurips_2025}

\usepackage[utf8]{inputenc} 
\usepackage[T1]{fontenc}    
\usepackage[pagebackref, breaklinks, colorlinks]{hyperref}       
\usepackage{url}            
\usepackage{booktabs}       
\usepackage{amsfonts}       
\usepackage{nicefrac}       
\usepackage{microtype}      
\usepackage{xcolor}         
\usepackage{multirow}
\usepackage{makecell}
\usepackage{graphicx}
\usepackage{pifont}
\usepackage{colortbl}
\usepackage{natbib}

\newlength\savewidth

\newcommand{\cmark}{\ding{51}}
\newcommand{\xmark}{\ding{55}}

\makeatletter

\newcommand{\Rmnum}[1]{\expandafter@slowromancap\romannumeral #1@}
\makeatother

\def\eg{\textit{e.g.}}

\title{Towards Unified Multimodal Interleaved Generation via Group Relative Policy Optimization}

\author{Ming Nie$^1$ \quad Chunwei Wang$^2$ \quad Jianhua Han$^3$ \quad Hang Xu$^3$ \quad Li Zhang$^1$\thanks{Li Zhang (lizhangfd@fudan.edu.cn) is the corresponding author.}
\vspace{.6em}
\\
$^1$School of Data Science, Fudan University \quad $^2$Noah's Ark Lab, Huawei
\\
$^3$Yinwang Intelligent Technology Co., Ltd.
\vspace{.6em}
\\
\url{https://github.com/LogosRoboticsGroup/UnifiedGRPO}
}

\begin{document}

\maketitle

\begin{abstract}
Unified vision-language models have made significant progress in multimodal understanding and generation, yet they largely fall short in producing multimodal interleaved outputs, which is a crucial capability for tasks like visual storytelling and step-by-step visual reasoning.
In this work, we propose a reinforcement learning-based post-training strategy to unlock this capability in existing unified models, without relying on large-scale multimodal interleaved datasets.
We begin with a warm-up stage using a hybrid dataset comprising curated interleaved sequences and limited data for multimodal understanding and text-to-image generation, which exposes the model to interleaved generation patterns while preserving its pretrained capabilities.
To further refine interleaved generation, we propose a unified policy optimization framework that extends Group Relative Policy Optimization (GRPO) to the multimodal setting.
Our approach jointly models text and image generation within a single decoding trajectory and optimizes it with our novel hybrid rewards covering textual relevance, visual-text alignment, and structural fidelity.
Additionally, we incorporate process-level rewards to provide step-wise guidance, enhancing training efficiency in complex multimodal tasks.
Experiments on MMIE and InterleavedBench demonstrate that our approach significantly enhances the quality and coherence of multimodal interleaved generation.
\end{abstract}
\section{Introduction}
In recent years, rapid progress in visual understanding and generation has driven a growing trend towards unifying these capabilities within a single multimodal framework.
Unified vision-language models aim to perform both understanding tasks (\eg, visual question answering) and generation tasks (\eg, text-to-image generation) within one model, representing a pivotal direction in the evolution of vision-language models.
This paradigm has seen significant advancement, enabled by the availability of large-scale image-text paired datasets and the scaling of multimodal model architectures.
Recent models such as Show-O, VILA-U, and ILLUME~\cite{xie2024show,wu2024vila,wang2024illume} demonstrate strong generalization across diverse benchmarks, underscoring the potential of unified frameworks as general-purpose AI systems for integrated vision-language reasoning and content creation.

Despite their impressive performance, most unified models still struggle to generate multimodal interleaved contents, which is a critical capability for enabling fine-grained reasoning, step-by-step explanation, and context-aware multimodal synthesis.
During inference, these models typically produce either text-only or image-only outputs, constrained by explicit or implicit modality control mechanisms.
This limitation arises primarily from the lack of fine-grained supervision and training data to guide dynamic modality transitions.
Consequently, current models often default to generating single-modality outputs, failing on tasks that require tightly coupled multimodal sequences, such as visual dialogue or sequential storytelling.
Addressing this gap is essential for realizing the full potential of unified models in seamless multimodal reasoning and generation.

To this end, we propose a post-training strategy that unlocks multimodal interleaved generation without requiring large-scale high-quality interleaved data.
We hypothesize that unified models inherently possess fundamental multimodal generation capabilities acquired during pre-training and supervised fine-tuning (SFT), and that minimal interleaved supervision is sufficient to activate them.
Based on this hypothesis, we first construct a hybrid dataset comprising a small amount of interleaved text-image sequences to expose the model to multimodal generation patterns, while incorporating limited SFT data to retain its pretrained strengths in multimodal understanding and conventional text-to-image generation.
After this warm-up stage, the unified model is capable of generating basic multimodal interleaved contents conditioned on instructions.
However, the outputs often suffer from weak cross-modal alignment, reflecting limited consistency and coherence between text and image.

To further improve multimodal interleaved generation, we propose a reinforcement learning algorithm that formulates generation as a sequential decision-making process and extends Group Relative Policy Optimization~\cite{shao2024deepseekmath} (GRPO) to the multimodal setting.
While GRPO has been widely explored for text-only modalities, it struggles with multimodal outputs due to challenges like modality switching and hybrid reward attribution.
To address this, we propose a unified policy optimization framework that jointly models text and image generation under a single decoding trajectory.
This approach enables modality-aware decisions and leverages a shared training objective, preserving GRPO's advantages while adapting to multimodal structures.
Specifically, we define a hybrid group-wise reward signal consisting of three key components:
(i) a textual reward that evaluates the quality and relevance of the generated text conditioned on the input prompt;
(ii) a visual and multimodal reward that jointly assesses image quality and image-text consistency;
and (iii) a format reward that promotes structural fidelity by penalizing violations in the expected interleaved output format.
In addition, we incorporate process-level reward modeling by assigning intermediate rewards at the interleaved generation steps.
This design provides more granular and timely feedback throughout the generation process, which significantly improves the efficiency and effectiveness of policy learning.
Such fine-grained supervision is particularly advantageous in complex multimodal generation tasks, where step-wise feedback helps the model perform autoregressive interleaved generation more effectively than relying solely on outcome rewards.
We evaluate our approach on two dedicated multimodal interleaved generation benchmarks, namely MMIE~\cite{xiammie} and InterleavedBench~\cite{Liu2024HolisticEF}, and demonstrate its effectiveness in enabling unified models to generate coherent, high-quality interleaved outputs.

To summarize, our main contributions are as follows:
\textbf{(i)}
We introduce a warm-up stage that unlocks the model’s latent capability for interleaved text-image generation using only a small amount of curated interleaved data;
\textbf{(ii)}
We propose a unified policy optimization framework that supports autoregressive generation across both text and image modalities, enabling seamless modality switching within a single decoding trajectory;
\textbf{(iii)}
We design a group of hybrid rewards that supervises multimodal generation from multiple aspects, and further enhance learning via process-level rewards that provide step-wise guidance;
\textbf{(iv)}
We conduct extensive experiments on two challenging benchmarks, MMIE and InterleavedBench, demonstrating the effectiveness and efficiency of our approach compared to existing unified models.

\section{Related work}
\subsection{Unified understanding and generation}
Recently, an increasing number of studies~\cite{xie2024show,zhou2024transfusion,sun2024emu2,team2024chameleon,wu2024vila,chen2025janus,wu2024liquid} have focused on unified models for both multimodal understanding and generation.
Show-o~\cite{xie2024show}, TransFusion~\cite{zhou2024transfusion} and Janus-Pro~\cite{chen2025janus} adopt hybrid diffusion-autoregressive strategies, while Emu2~\cite{sun2024emu2} employs a fully autoregressive architecture that predicts the next multimodal element via classification for text and regression for visual embeddings.
Chameleon~\cite{team2024chameleon}, VILA-U~\cite{wu2024vila}, and other works~\cite{wang2024illume,wu2024liquid} tokenize images and interleave them with text tokens, enabling joint reasoning and autoregressive generation across modalities.
Despite these efforts, current unified models often fail to support fine-grained interleaved generation of text and images, largely due to the scarcity of high-quality multimodal interleaved data.
This limitation restricts their applicability in tasks that require coherent and context-aware multimodal dialogue and reasoning.

\subsection{Multimodal interleaved generation}
Multimodal learning has rapidly advanced, especially in the integration of text and image modalities.
Early models such as DALL·E~\cite{ramesh2021dalle} and Stable Diffusion~\cite{podell2023sdxl} demonstrated impressive capabilities in generating images from text prompts, but they were primarily designed for unidirectional generation, either text-to-image or image-to-text, without supporting interleaved multimodal outputs.
Recently, large vision-language models (LVLMs) have begun to tackle this limitation by enabling interleaved generation, where text and images are jointly generated within a single autoregressive sequence.
Approaches such as Chameleon~\cite{team2024chameleon}, GILL~\cite{gill}, Anole~\cite{chern2024anole} and ARMOR~\cite{sun2025armor}, have pushed the boundaries of token-level multimodal modeling.
These models adopt unified generation frameworks that allow seamless alternation between modalities, supporting more natural and interactive multimodal communication.
However, a key challenge lies in the scarcity of large-scale, fine-grained multimodal supervision data, which limits the full potential of interleaved generation.
To address this, we propose a post-training strategy that activates the model’s multimodal generation capability without relying on extensive high-quality training data.

\subsection{Policy optimization in multimodal models}
Policy optimization~\cite{ouyang2022rlhf,dpo,shao2024deepseekmath,guo2025deepseek,shen2025vlmr1,wei2025skywork} has become a pivotal component in aligning large language and vision-language models with human intent.
Reinforcement learning from human feedback (RLHF)~\cite{ouyang2022rlhf}, exemplified by methods such as PPO~\cite{schulman2017proximal} and DPO~\cite{dpo}, has proven effective in improving text-only generation by leveraging reward signals derived from preference data or learned reward models.
Recently, Group Relative Policy Optimization (GRPO)~\cite{shao2024deepseekmath} has demonstrated promising results in the post-training of large language models, particularly in domains requiring complex reasoning such as mathematics~\cite{guo2025deepseek,shen2025vlmr1,wei2025skywork}.
Its group-wise comparative reward mechanism offers improved sample efficiency and stability, making it a compelling foundation for extension into multimodal interleaved generation.
However, extending policy optimization to multimodal settings remains a substantial challenge.
First, existing methods~\cite{zhou2025r1,wang2025multimodalcot,chen2025towards} are often limited to text-only modalities or apply separate optimization procedures for text and image components, which hinders the development of unified models capable of seamless interleaved generation.
Second, most approaches rely on outcome-oriented, end-of-sequence rewards, which are inherently sparse and fail to capture the fine-grained, step-wise alignment required for complex multimodal tasks.
In this work, we propose a unified policy optimization algorithm, which treats multimodal generation as a single coherent decision-making process for enhanced multimodal policy optimization.

\section{Methodology}
In this section, we first present the preliminary formulation of unified multimodal models in ~\ref{preliminary}.
We then introduce a warm-up training scheme to equip the model with the capability of generating interleaved text-image outputs while preserving its pretrained knowledge in Sec.~\ref{warmup}.
Next, in Sec.~\ref{grpo}, we describe our proposed GRPO training framework based on hybrid reward signals tailored for multimodal generation.
Finally, we outline the training details in Sec.~\ref{details}.

\subsection{Preliminary}\label{preliminary}
Unified multimodal models seek to integrate vision and language by leveraging a shared architecture capable of jointly modeling textual and visual modalities in both understanding and generation tasks.
Given a multimodal input sequence $X = \{x_1, x_2, \ldots, x_n\}$, where each token $x_i$ can correspond to either a textual or visual token, the model learns to autoregressively generate an output sequence $Y = \{y_1, \ldots, y_m\}$, which is also capable of comprising both modalities.
The generation process in unified models is typically modeled as an autoregressive probability distribution over the output tokens:

\begin{equation}
    p(Y|X) = \prod_{t=1}^{m}p(y_t|X,y_{<t};\theta),
\end{equation}

where $\theta$ denotes the model parameters, and the conditional generation at each step relies on both the input sequence $X$ and the previously generated tokens $y_{<t}$.
The unified model is generally optimized using a standard language modeling objective covering multimodal sequences:

\begin{equation}
    \mathcal{L}=-(\sum_{t=1}^{T}\mathbb{I}_{txt}(t)logP(y_t|x,y_{<t})+\sum_{t=1}^{T}\mathbb{I}_{img}(t)logP(y_t|x,y_{<t})).
\end{equation}

This framework allows for flexible training across various tasks simultaneously, such as captioning, image generation, and more complex multimodal reasoning.
However, due to the lack of high-quality training data that explicitly supervises interleaved text-image generation, current unified models often struggle to realize their potential for multimodal generation, leading to limited and inconsistent performance in such tasks.

\begin{figure*}[t]
    \begin{center}
        \includegraphics[width=1.0\linewidth]{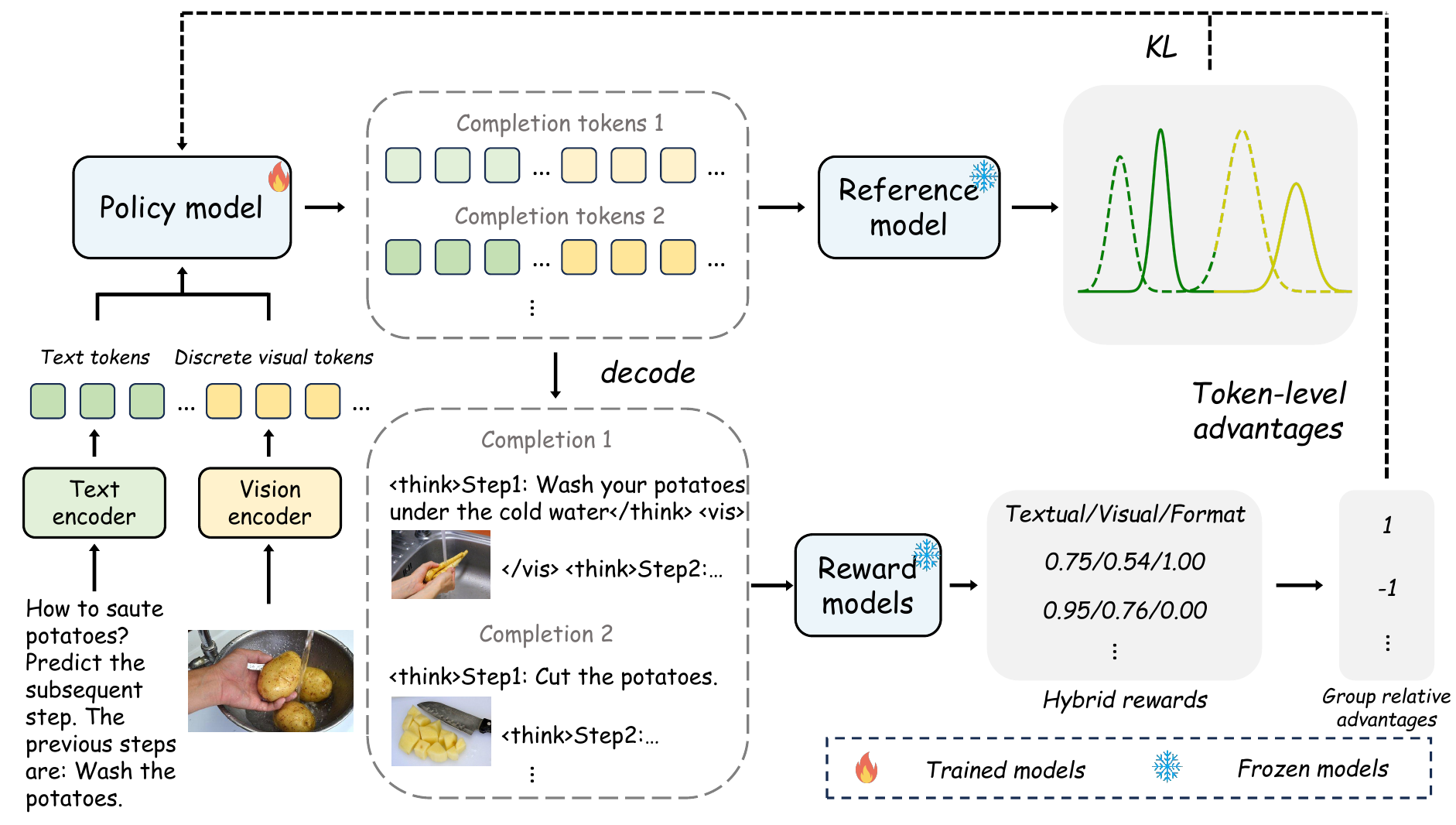}
    \end{center}
    \caption{Overview of our reinforcement fine-tuning framework.
    Multimodal tokens are autoregressively generated and decoded into completions, with token probabilities used to compute KL divergence along a single trajectory.
    Hybrid rewards are assigned to each completion, and token-level group relative advantages are calculated to guide policy optimization along with KL regularization.}
    \label{fig-method}
\end{figure*}

\subsection{Unlocking interleaved generation}\label{warmup}
To equip unified models with interleaved generation and multimodal reasoning capabilities, we propose a warm-up training scheme that activates new functionalities while preserving existing strengths.
Unified models are pretrained on large-scale multimodal data comprising both text and images, theoretically equipping them for multimodal interleaved generation.
However, they often struggle with smoothly transitioning between modalities and aligning context across text and image.
Direct fine-tuning on novel tasks may substantially impair the model’s pretrained competencies and induce catastrophic forgetting.
To address this, we introduce a hybrid data-based warm-up phase that bridges pretraining and task-specific fine-tuning.
In this stage, we incorporate a large corpus of high-quality supervised fine-tuning (SFT) data, including either curated human-labeled samples or distilled outputs from strong multimodal LLMs.
We further introduce interleaved text-image data to expose the model to multimodal generation patterns, while blending in purely textual data to preserve and reinforce its language understanding and generation capabilities.
This strategy helps unlock interleaved generation abilities without compromising the model’s original strengths.

\subsection{Reinforcement fine-tuning}\label{grpo}
Once the model is warmed up, we proceed to a GRPO-based optimization phase to further enhance generation quality and cross-modal alignment.
During reinforcement fine-tuning, we apply the GRPO algorithm guided by a hybrid supervised signal comprising both rule-based metrics and model-based reward, as shown in Figure~\ref{fig-method}.

\noindent\textbf{Unified policy for multimodal generation.}
To enhance interleaved text-image generation and multimodal reasoning, we adapt Group Relative Policy Optimization (GRPO)~\cite{shao2024deepseekmath}, an efficient reinforcement learning algorithm originally proposed for text-only LLMs.
GRPO estimates token-level advantages by conducting intra-group comparisons among generated responses conditioned on the same query.
Given an input instance $X$, a behavior agent $\phi_{\theta_{old}}$ samples a batch of $G$ candidate responses $\{Y_i\}^{G}_{i=1}$.
However, previous applications of GRPO have been limited to optimizing text-only policies.
To extend it to unified models capable of generating both text and images, we treat the multimodal generation as a single decision process: $Y_i=\{y_1^{txt},\ldots,y_{k}^{txt},y_{k+1}^{img},\ldots,y_{m}^{img}\}$, and adapt the GRPO framework accordingly.
The GRPO optimization objective incorporates a clipped surrogate loss term augmented with a KL-penalty to ensure stable policy updates, formulated as:

\begin{equation}
    \mathcal{L}_{GRPO}(\theta)=\frac{1}{G}\sum^{G}_{i=1}\frac{1}{|Y_i|}\sum^{|Y_i|}_{t=1}min[\mathbb{D}_{uni}(t)\hat{A}_{i,t}, clip(\mathbb{D}_{uni}(t), 1-\epsilon,1+\epsilon)\hat{A}_{i,t}],
\end{equation}

where $\mathbb{D}_{uni}$ represents the unified KL divergence computed per token across modalities:

\begin{equation}
    \mathbb{D}_{uni}(t)=\frac{\pi_{\theta}(y_{i,t}^{txt}|X_i,y_{i,\leq t})}{\pi_{\theta_{old}}(y_{i,t}^{txt}|X_i,y_{i,\leq t})}\mathbb{I}_{t\leq k}(t) + \frac{\pi_{\theta}(y_{i,t}^{img}|X_i,y_{i,\leq t})}{\pi_{\theta_{old}}(y_{i,t}^{img}|X_i,y_{i,\leq t})}\mathbb{I}_{t>k}(t).
\end{equation}

The advantage $\hat{A}_{i,t}$ for the $i$-th response at time step $t$ is determined by normalizing the rewards obtained across the response group, according to:

\begin{equation}
    \hat{A}_{i,t} = \frac{r(X,Y_i) - mean(\{r(X,Y_1),...,r(X,Y_G)\})}{std(\{r(X,Y_1),...,r(X,Y_G)\})}.
\end{equation}

The hyperparameter $\epsilon$ controls the permissible deviation from the reference policy.
The clipping function constrains the policy ratio within a specified interval, preventing overly large updates.
This regularization mechanism enhances training stability and reduces the risk of performance collapse due to aggressive policy shifts.

\noindent\textbf{Hybrid reward signal.}
Formally, a group of reward models assign a scalar score to each response, producing a set of $G$ rewards $r = \{r_1, \ldots, r_G\}$ corresponding to the $G$ candidates.
To guide the optimization of the unified model, we design a hybrid reward signal that integrates multiple task-specific objectives.
The hybrid reward aims to jointly enhance the quality of both generated text and images, as well as ensure output format consistency.

To better model and optimize the interleaved multimodal outputs, we begin with a probabilistic decomposition of the joint distribution over input prompt $X$, generated text $Y_{txt}$, and generated image $Y_{img}$.
We decompose the joint probability as:

\begin{equation}
    p(Y_{txt},Y_{img}|X) = p(Y_{txt}|X)p(Y_{img}|X,Y_{txt}).
\end{equation}

From this perspective, we design our hybrid reward to align with the two conditional components:
The textual reward $r_t$ evaluates $p(Y_{txt}|X)$, assessing the relevance and coherence of the generated text given the prompts.
The visual rewards $r_v$ targets $p(Y_{img}|X,Y_{txt})$, measuring the image's quality and its alignment with the text and prompt context.
Furthermore, we incorporate a format reward $r_f$ to regularize the structure of the generated content.
Specifically, we leverage special tokens \texttt{<think>} and \texttt{<vis>} to explicitly separate different modalities in the content sequence, guiding the model to adhere to a consistent and interpretable format.
This formatting strategy is illustrated in Figure~\ref{fig-method}.
The textual relevance reward $r_t$ is assessed by~\cite{xiong2025llavacritic} and the visual reward is assessed by ImageReward~\cite{xu2023imagereward}.
To this end, our hybrid reward function is defined as follows:

\begin{equation}
    r(X,Y_i)=r_t(X,Y_i)+r_v(X,Y_i)+r_f(X,Y_i).
\end{equation}

\noindent\textbf{Process-level reward.}
While outcome supervision provides a single reward at the end of each output, this sparse feedback may be insufficient and inefficient for complex multimodal generation tasks.
To address this, we additionally incorporate process-level supervision, which offers intermediate rewards at the end of each modality step.
Formally, given a set of sampled outputs $\{Y_i\}_{i=1}^{G}$, a process reward model assigns a sequence of rewards to each output: $R=\{\{r_1^{index(1)},...,r_1^{index(K_1)}\},...,\{r_G^{index(1)},...,r_G^{index(K_G)}\}\}$, where $index(j)$ denotes the end token index of the $j$-th step, and $K_i$ is the total number of modality steps in the $i$-th output.
Each step reward is then normalized using the overall mean and standard deviation across all samples.
To compute token-level advantages, we accumulate the normalized rewards from all subsequent steps: $\hat{A}_{i,t}=\sum_{index(j)\geq t}\hat{r}_i^{index(j)}$.
The final policy is then optimized by maximizing the GRPO objective using these token-level advantages.

\subsection{Training details}\label{details}
We implement VILA-U~\cite{wu2024vila} as our foundation unified model.
Thanks to its unified pretraining paradigm, which jointly learns multimodal comprehension and text-to-image generation, the model inherently possesses the potential for multimodal output.
Our approach builds on this capability and unlocks interleaved generation with only minimal additional data.
During the warm-up stage, we collect 0.3M text-image paired samples with interleaved outputs from ActivityNet~\cite{caba2015activitynet}, GenHowTo~\cite{souvcek2024genhowto} and OpenStory++~\cite{ye2024openstory++}.
To preserve the model’s original multimodal understanding and generation abilities, we further incorporate 1M multimodal understanding samples from EMOVA~\cite{chen2024emova} and 1M text-to-image generation samples from JourneyDB~\cite{sun2023journeydb}.
For the GRPO stage, we curate a dataset of 0.1M samples (from the same sources as in the warm-up stage) focusing on visual storytelling and multimodal interleaved reasoning to facilitate effective policy optimization.

\section{Experiment}\label{exp}
\begin{table}[t]
\renewcommand\arraystretch{1.5}
\begin{center}
\resizebox{1.0\linewidth}{!}{
\begin{tabular}{c|c|cccc}
\toprule[1.5pt]
& & \multicolumn{4}{c}{MMIE} \\
\multirow{-2}{*}{Method} & \multirow{-2}{*}{Params} & Situational analysis & Project-based learning & Multi-step reasoning & AVG \\
\midrule[1.0pt]
MiniGPT-5~\cite{zheng2023minigpt} & 7B & 47.63 & 55.12 & 42.17 & 50.92 \\
EMU-2~\cite{sun2024emu2} & 37B & 39.65 & 46.12 & 50.75 & 45.33 \\
GILL~\cite{gill} & 7B & 46.72 & 57.57 & 39.33 & 51.58 \\
Anole~\cite{chern2024anole} & 7B & 48.95 & 59.05 & 51.72 & 55.22 \\
\midrule
Ours (QLoRA) & 7B & 53.12 & 60.34 & 53.28 & 57.27 \\ 
Ours & 7B & \textbf{56.87} & \textbf{62.28} & \textbf{54.31} & \textbf{59.50} \\
\bottomrule[1.5pt]
\end{tabular}}
\end{center}
\vspace{3mm}
\caption{Comparison with existing methods on MMIE.
The proposed method significantly improves the generation quality of unified models, producing interleaved text–image outputs that are both coherent and aligned with instructions.
}
\label{tab-mmie}
\end{table}
\begin{table}[t]
\renewcommand\arraystretch{1.5}
\begin{center}
\resizebox{0.9\linewidth}{!}{
\begin{tabular}{c|c|cccccc}
\toprule[1.5pt]
& & \multicolumn{6}{c}{InterleavedBench} \\
\multirow{-2}{*}{Method} & \multirow{-2}{*}{Params} & Text quality & Perceptual quality & Image coherence & TIC & Helpfulness & AVG \\
\midrule[1.0pt]
MiniGPT-5~\cite{zheng2023minigpt} & 7B & 1.22 & 2.45 & 1.62 & 2.03 & 1.77 & 1.82 \\
EMU-2~\cite{sun2024emu2} & 37B & 1.26 & 2.28 & 1.89 & 1.34 & 1.64 & 1.68 \\
GILL~\cite{gill} & 7B & 0.75 & 3.21 & 2.25 & 1.53 & 1.48 & 1.84 \\
\midrule
Ours (QLoRA) & 7B & 2.33 & 3.39 & 3.01 & 2.75 & 2.85 & 2.87 \\ 
Ours & 7B & \textbf{2.86} & \textbf{3.58} & \textbf{3.25} & \textbf{3.02} & \textbf{2.94} & \textbf{3.13} \\
\bottomrule[1.5pt]
\end{tabular}}
\end{center}
\vspace{3mm}
\caption{Comparison with existing methods on InterleavedBench, where our method demonstrates superior performance across multiple evaluation metrics.}
\label{tab-ib}
\end{table}
\subsection{Experimental setup}
\noindent\textbf{Data preparation.}
During the warm-up stage, we collect 0.3M interleaved text-image samples from ActivityNet~\cite{caba2015activitynet}, GenHowTo~\cite{souvcek2024genhowto}, and OpenStory++~\cite{ye2024openstory++}.
The data covers diverse multimodal scenarios, including temporally grounded actions, step-by-step state transitions, and narrative storytelling, providing rich interleaved patterns for training.
\textbf{(i)}
ActivityNet~\cite{caba2015activitynet} is a large-scale benchmark for human activity understanding, covering 203 activity classes with about 137 untrimmed videos per class and 1.41 activity instances per video (849 hours total).
We use its temporal dense captions~\cite{krishna2017dense} to describe fine-grained action segments.
For each localized clip, a duration-adaptive number of keyframes is extracted to capture visual details, and these keyframes with their captions form interleaved training samples.
\textbf{(ii)}
GenHowTo is a large-scale dataset automatically constructed with 200K image triplets and corresponding textual descriptions.
Each triplet contains temporally ordered frames depicting an object’s initial state, the modifying action, and the resulting new state, effectively modeling cause-and-effect visual transitions.
These structured triplets naturally support interleaved text-image training for step-by-step visual reasoning.
We leverage both the provided action and state prompts, along with their aligned images, to construct coherent interleaved text-image samples for visual reasoning tasks.
\textbf{(iii)}
OpenStory++~\cite{ye2024openstory++} is a comprehensive dataset emphasizing narrative continuity with instance-level visual segmentation.
It extracts keyframes from videos, filters them by aesthetic quality, and generates refined captions via an LLM-based annotator to ensure coherence.
The resulting keyframe–caption pairs form interleaved text-image sequences for visual storytelling.
More detailed data construction process and illustration are provided in the appendix.

\noindent\textbf{Implementation details.}
As described previously, we adopt VILA-U~\cite{wu2024vila} as our foundation unified model.
The language model, vision encoder, and visual decoder are all implemented following VILA-U's original architecture and configuration.
All images are resized to a fixed resolution of 256 $\times$ 256 before being fed into the model.
For image generation, we employ classifier-free guidance with a guidance scale of 3 to improve output fidelity.
In GRPO stage, the number of generation $G$ is set to 4 and we train the model for 3k steps.
All full-scale experiments are conducted using 32 NVIDIA A100 GPUs.
To further demonstrate the compatibility and efficiency of our method, we also implement a lightweight version using QLoRA, achieving competitive performance with significantly lower memory and compute requirements.

\noindent\textbf{Evaluation.}
We evaluate the model’s capability for interleaved multimodal generation on two dedicated benchmarks: MMIE\cite{xiammie} and InterleavedBench\cite{Liu2024HolisticEF}.
MMIE consists of 20K carefully curated multimodal queries, covering 3 main categories, 12 domains, and 102 subfields, including mathematics, coding, physics, literature, health, and the arts.
It introduces a robust automated evaluation protocol using a scoring model fine-tuned on human-annotated data with systematic criteria, aiming to reduce bias and enhance evaluation reliability.
InterleavedBench comprises 815 instances across 10 real-world use cases.
It defines five key evaluation dimensions: text quality, perceptual quality, image coherence, text-image coherence, and overall helpfulness, offering a comprehensive and fine-grained assessment of interleaved generation performance.
By encompassing detailed evaluation criteria, InterleavedBench provides a robust framework for measuring a model’s ability to generate coherent, visually aligned, and contextually meaningful interleaved outputs.

\subsection{Main results}
We compare our approach against existing unified models on these challenging benchmarks to evaluate its effectiveness in interleaved multimodal generation.
As shown in Table~\ref{tab-mmie}, our approach significantly improves the ability of unified models to produce coherent and instruction-aligned text-image outputs.
These improvements demonstrate that our method not only strengthens the generation quality under limited supervision, but also pushes the performance upper bound of current unified models on interleaved multimodal generation benchmarks (59.5\% on MMIE).
In particular, our method demonstrates a clear advantage on the situational analysis task, which evaluates a model's ability to perform visual storytelling based on given textual and visual prompts.
Our model achieves a score of 56.87\%, outperforming Anole by over 10\%.
This advanced performance highlights the effectiveness of our interleaved generation strategy in handling complex, context-rich multimodal reasoning scenarios.
We also report our results on InterleavedBench, as shown in Table~\ref{tab-ib}.
Our method outperforms existing unified models, achieving superior performance with a gain of 1.29 over GILL, demonstrating the robustness and generalization ability of our approach.

\begin{table}[t]
\renewcommand\arraystretch{1.5}
\begin{center}
\begin{tabular}{cc|cc}
\toprule[1.5pt]
Warm-up & GRPO & MMIE & InterleavedBench \\
\midrule[1.0pt]
\xmark & \xmark & - & 0.51 \\
\cmark & \xmark & 53.31 & 1.97 \\
\rowcolor[gray]{0.9}\cmark & \cmark & 59.50 & 3.13 \\
\bottomrule[1.5pt]
\end{tabular}
\end{center}
\vspace{3mm}
\caption{Effects of warm-up stage and GRPO.
The warm-up stage enables to generate interleaved multimodal outputs.
Building on the warm-up initialization, the proposed GRPO-based policy optimization substantially improves the quality of interleaved generations.
}
\label{tab-ab1}
\end{table}

\begin{table}[t]
\renewcommand\arraystretch{1.5}
\begin{center}
\begin{tabular}{cccc|cc}
\toprule[1.5pt]
\multicolumn{3}{c}{Hybrid rewards} & & & \\
$r_f$ & $r_t$ & $r_v$ & \multirow{-2}{*}{Process reward} & \multirow{-2}{*}{MMIE} & \multirow{-2}{*}{InterleavedBench} \\
\midrule[1.0pt]
\cmark & \xmark & \xmark & \xmark & 53.56 & 2.05 \\
\cmark & \cmark & \xmark & \xmark & 54.62 & 2.30\\
\cmark & \cmark & \cmark & \xmark & 57.83 & 2.79 \\
\rowcolor[gray]{0.9}\cmark & \cmark & \cmark & \cmark & 59.50 & 3.13 \\
\bottomrule[1.5pt]
\end{tabular}
\end{center}
\vspace{3mm}
\caption{Hybrid reward components ablation.
Beginning with the format reward $r_f$, we gradually integrate the textual, visual and process-level rewards.
The results demonstrate that the inclusion of complementary reward signals progressively enhances interleaved generation quality.
}
\label{tab-ab2}
\end{table}

\subsection{Ablation study}
\noindent\textbf{Effects of warm-up stage and GRPO.}
We first investigate the impact of GRPO as well as warm-up stage in Table~\ref{tab-ab1}.
Empirical results demonstrate that the warm-up stage effectively unlocks the unified model’s capability to generate interleaved multimodal outputs (53.31\% on MMIE and 1.97 on InterleavedBench).
Prior to this stage, existing unified models were unable to produce such interleaved generations, and thus failed to yield valid results on benchmarks like MMIE.
Notably, InterleavedBench includes a subset of tasks that do not require multimodal outputs, which allows baseline models to achieve a marginal score (0.51).
Building upon the warm-up initialization, our proposed GRPO-based policy optimization significantly enhances the quality of interleaved outputs, leading to consistent improvements across both benchmarks (+6.19\% and +1.16).
This two-stage strategy proves essential for enabling instruction-following, coherent, and structurally aligned text-image sequences.

\noindent\textbf{Rewards components ablation.}
We then take a further step to investigating the components of rewards in the GRPO training paradigm.
Starting from the format reward $r_f$, we progressively incorporate the textual reward $r_t$, visual reward $r_v$, and finally the process-level reward.
As shown in Table~\ref{tab-ab2}, the format reward alone does not lead to significant performance gains compared to the warm-up stage.
However, it plays an essential role in providing explicit supervision for modality switching, which is crucial for stable interleaved generation.
Adding the textual reward yields a steady improvement of 1.06\% on MMIE and 0.25 on InterleavedBench, while the visual reward further boosts performance by 3.01\% and 0.49, respectively.
Finally, integrating the process-level reward results in the best performance, raising the scores to 59.50\% on MMIE and 3.13 on InterleavedBench.
These results demonstrate the effectiveness of each component and highlight the importance of fine-grained, multi-dimensional reward signals in training high-quality interleaved generation models.

\begin{table}[t]
\renewcommand\arraystretch{1.5}
\begin{center}
\resizebox{1.0\linewidth}{!}{
\begin{tabular}{c|c|c|cccc|ccc}
\toprule[1.5pt]
& & & \multicolumn{4}{c|}{Understanding} & \multicolumn{3}{c}{Generation} \\
\multirow{-2}{*}{Method} & \multirow{-2}{*}{ILO-Uni.} & \multirow{-2}{*}{Params} & MME-P~\cite{liang2024mme} & MMvet~\cite{yu2024mmvet} & SEEDBench-img~\cite{li2024seedbench} & POPE~\cite{li2023pope} & \#Train images & Image res. & Score \\
\midrule[1.0pt]
VILA-U~\cite{wu2024vila} & \xmark & 7B & 1401.8 & 33.5 & 59.0 & 85.8 & 15M & 384 & 0.42 \\
SEED-X~\cite{ge2024seed} & \xmark & 17B & 1435.7 & - & - & 84.2 & 158M & - & 0.49 \\
TokenFlow-XL~\cite{qu2024tokenflow} & \xmark & 13B & 1545.9 & 40.7 & 68.7 & 86.8 & 60M & 384 & 0.55 \\
Janus-Pro~\cite{chen2025janus} & \xmark & 7B & 1516.7 & 45.1 & 70.1 & 78.9 & 72M & 384 & 0.80 \\
\midrule
Chameleon~\cite{team2024chameleon} & \cmark & 7B & 153.1 & 8.3 & 30.5 & 19.4 & 1.4B & 512 & 0.39 \\
ARMOR~\cite{sun2025armor} & \cmark & 8B & 1619.4 & 53.1 &  74.8 &  87.7 & 5M & 256 & 0.37 \\
Ours & \cmark & 7B & 1425.2 & 32.8 & 59.2 & 85.1 & 16M & 256 & 0.46 \\
\bottomrule[1.5pt]
\end{tabular}}
\end{center}
\vspace{3mm}
\caption{Comparison with existing methods on visual understanding and generation benchmarks.
ILO-Uni. denotes ``Unified models with interleaved text-image output''.
Our approach maintains comparable performance to baseline models.
}
\label{tab-compgen}
\end{table}
\begin{table}[t]
\renewcommand\arraystretch{1.5}
\begin{center}
\begin{tabular}{c|c|c|cc}
\toprule[1.5pt]
KL & $G$ & $r_v$ & MMIE & InterleavedBench\\
\midrule[1.0pt]
\xmark & 2 & ImageReward~\cite{xu2023imagereward} & 53.04 & 1.72 \\
\cmark & 2 & ImageReward & 55.14 & 2.27 \\
\rowcolor[gray]{0.9}\cmark & 4 & ImageReward & 59.50 & 3.13 \\
\cmark & 4 & CLIP-score & 56.94 & 2.58 \\
\bottomrule[1.5pt]
\end{tabular}
\end{center}
\vspace{3mm}
\caption{Aspects of GRPO hyper-parameter implementation.
We first analyze the effect of the KL-penalty term by duplicating its application during training.
Results show that maintaining a KL constraint stabilizes optimization and mitigates policy drift from the initialization, underscoring its critical role.
We then ablate the number of generations $G$ to evaluate its impact on learning dynamics, and finally, we compare different formulations of the visual reward function.
}
\label{tab-ab3}
\end{table}

\noindent\textbf{Unified understanding and generation.}
Additionally, we evaluate our model's performance on standard multimodal understanding and generation tasks (GenEval~\cite{ghoshgeneval}) to ensure that our method does not lead to performance degradation or catastrophic forgetting of existing unified capabilities.
As shown in Table~\ref{tab-compgen}, our approach achieves comparable results to the baseline model across these tasks, indicating that both the warm-up stage and our GRPO-based training strategy are effective in preserving the model’s original strengths.
Furthermore, we compare our method with existing unified models that are capable of generating multimodal interleaved outputs.
Results show that our approach achieves competitive performance, highlighting its advantage in enabling high-quality and instruction-aligned interleaved generation without sacrificing general multimodal capabilities.

\noindent\textbf{Aspects of GRPO hyper-parameter implementation.}
We conduct a series of ablation studies to examine the impact of key hyper-parameter choices in our GRPO framework, as shown in Table~\ref{tab-ab3}.
We begin by analyzing the role of the KL-penalty term, where we duplicate its application during training.
Results show that enforcing KL divergence stabilizes training and prevents the model from drifting too far from the initial policy, confirming its importance.

We further ablate the number of generations $G$ for each prompt during training.
Increasing the number of sampled generations from 2 to 4 results in a notable performance improvement (+4.36\% and +0.86), indicating that generating a more diverse set of candidates facilitates more accurate group-wise reward estimation and provides richer learning signals.
However, due to computational limitations, we are unable to explore larger generation numbers.
It is worth noting that GRPO is particularly resource-intensive in the multimodal setting, as it requires generating both textual and visual outputs during each rollout.
This underscores the importance of developing more efficient and scalable post-training strategies for multimodal models in future research.

Lastly, we compare different choices of visual reward functions $r_v$.
Specifically, we find that using ImageReward provides more accurate and consistent supervision signals for image quality than CLIP-score, reinforcing the value of tailored reward models in multimodal training.

\begin{figure*}[t]
    \begin{center}
        \includegraphics[width=1.0\linewidth]{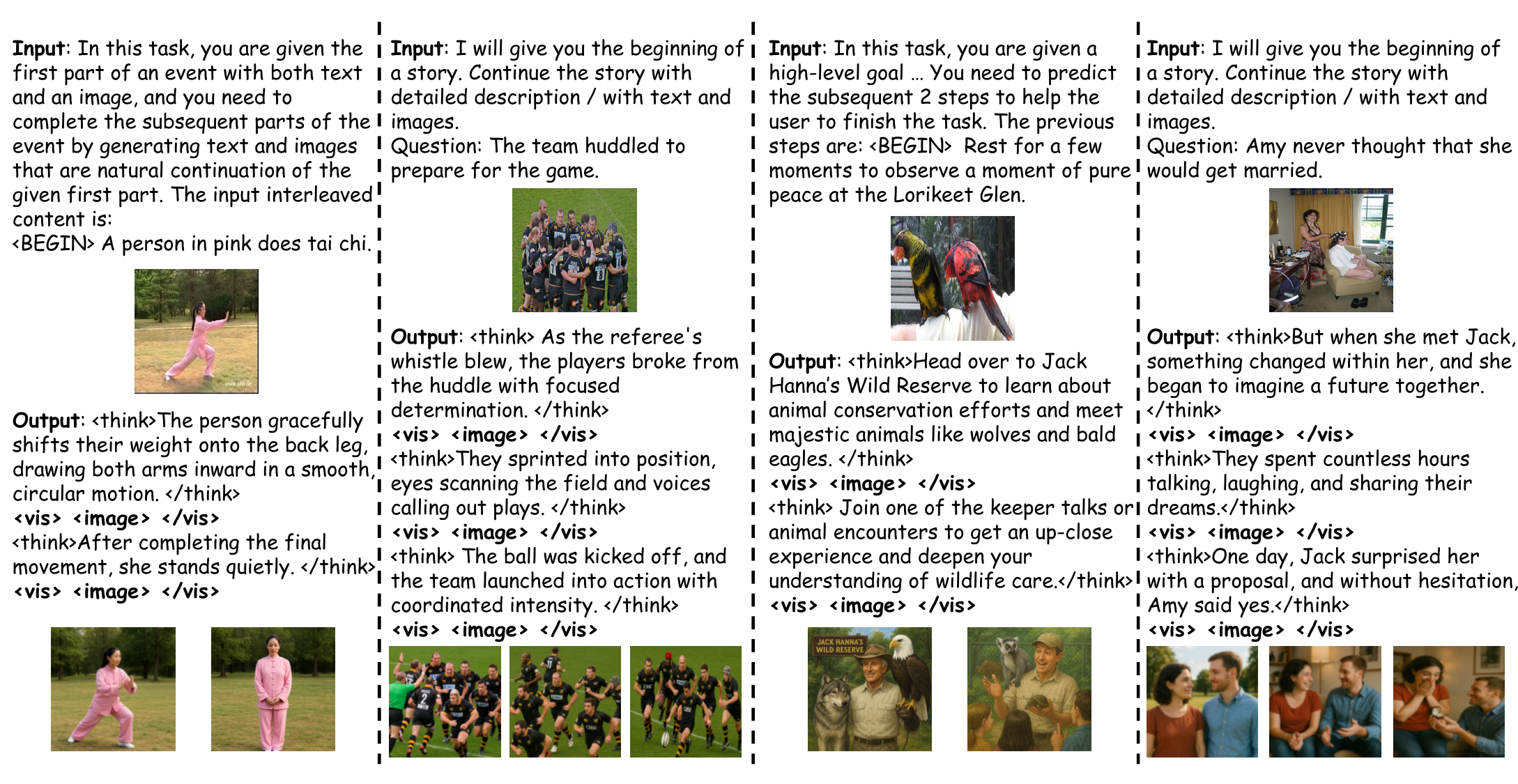}
    \end{center}
    \caption{Visualizations of multimodal interleaved generation.
    Qualitative examples illustrate the model’s capacity to produce coherent interleaved outputs, smoothly transitioning between text and image modalities within a unified generation process.}
    \label{fig-vis}
\end{figure*}

\noindent\textbf{Visualizations.}
Figure~\ref{fig-vis} showcases qualitative examples of interleaved generation, illustrating the model’s ability to seamlessly alternate between text and image modalities.
Notably, the visual outputs align well with the surrounding textual content, reflecting the effectiveness of our policy optimization strategy in producing fine-grained, semantically consistent interleaved content.
More visualizations and failure case studies are included in the appendix.

\section{Conclusion and discussions}\label{concl}
\noindent\textbf{Conclusion.}
We propose an effective training strategy to enable unified vision-language models to perform high-quality interleaved multimodal generation.
Through a warm-up stage with limited data, we unlock the model’s latent multimodal generation capabilities.
We further introduce a unified policy optimization framework based on GRPO with hybrid and process-level rewards, which formulates the multimodal generation process as a single trajectory from a reinforcement learning perspective.
Experiments on MMIE and InterleavedBench demonstrate superior performance over existing unified models while preserving general multimodal capabilities, paving the way for more versatile and controllable multimodal generation systems.

\noindent\textbf{Discussion on limitations.}
Despite the significant performance gains on multimodal interleaved benchmarks, we observe that our method does not bring notable improvements on generalized multimodal understanding and text-to-image generation tasks.
We attribute this to the underlying base model, which largely determines performance limits, with GRPO mainly serving to better align text and image content rather than enhancing general capabilities.
Future work may benefit from stronger unified architectures and broader reward designs to further enhance model capability.

\section*{Acknowledgments}
This work was supported in part by National Natural Science Foundation of China (Grant No. 62376060).

\bibliographystyle{plain}
\bibliography{egbib}

\newpage
\appendix
\section{Data preparation}
During the warm-up stage, we collect 0.3M text-image paired samples with interleaved outputs from ActivityNet~\cite{caba2015activitynet}, GenHowTo~\cite{souvcek2024genhowto} and OpenStory++~\cite{ye2024openstory++}.
The data comprises diverse multimodal scenarios, including temporally grounded action descriptions, step-by-step action-state pairs, and narrative visual storytelling.
These samples are organized into interleaved text-image sequences to expose the model to rich multimodal generation patterns during training.

\noindent\textbf{ActivityNet.}
ActivityNet~\cite{caba2015activitynet} is a large-scale video benchmark designed for human activity understanding.
It covers a broad spectrum of complex daily activities, featuring 203 activity classes with an average of 137 untrimmed videos per class and 1.41 activity instances per video, totaling approximately 849 hours of video content.
We utilize the temporal dense captions that describe fine-grained action segments in each video~\cite{krishna2017dense}.
For each temporally localized clip, we extract a set of representative keyframes to capture fine-grained visual details.
The number of keyframes is adaptively determined based on the duration of the action clip.
These keyframes, along with their corresponding dense captions, are subsequently used to construct interleaved training samples.

\begin{figure*}[h]
    \begin{center}
        \includegraphics[width=1.0\linewidth]{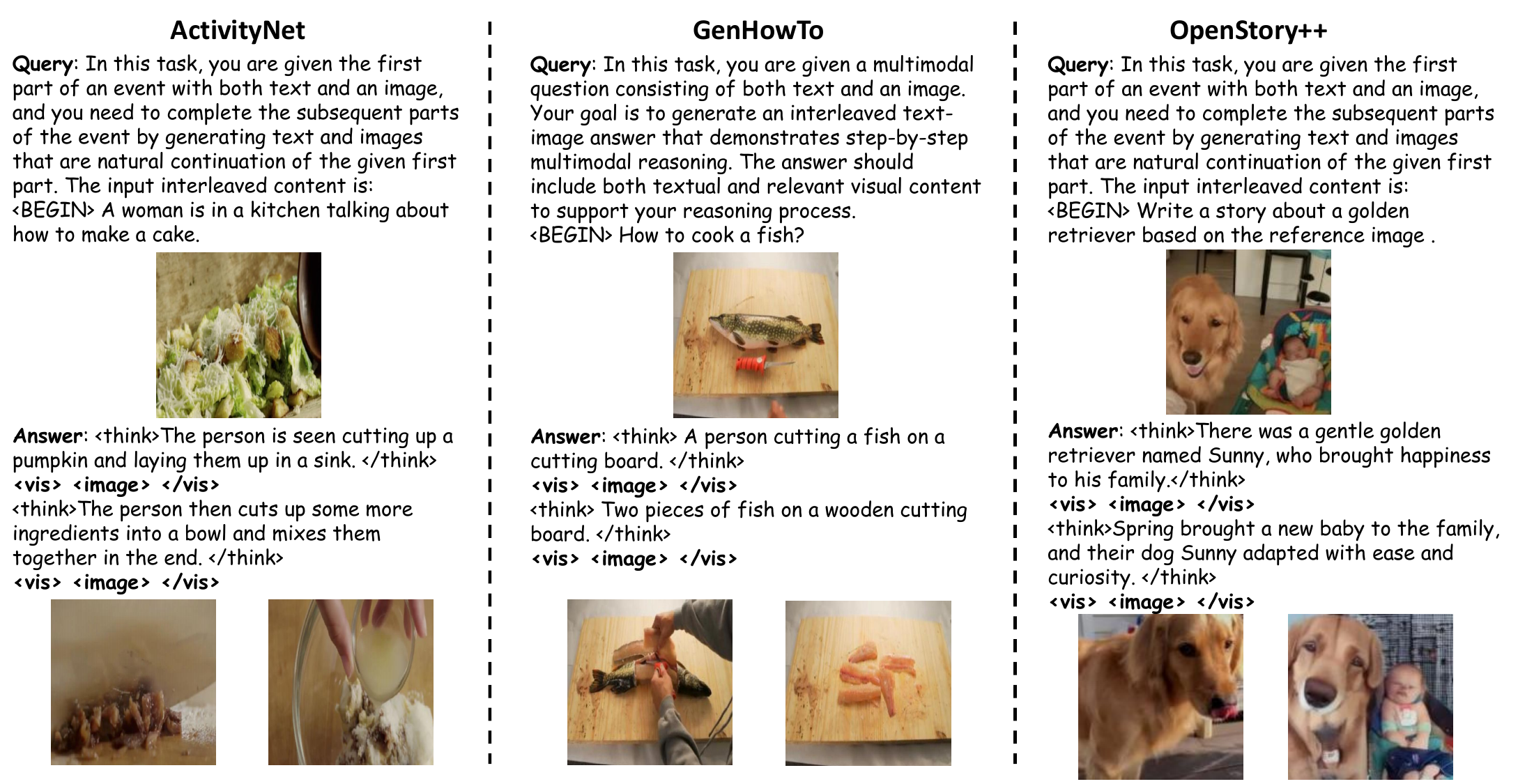}
    \end{center}
    \caption{Illustration of data preparation process during the warm-up stage.}
    \label{fig-data}
\end{figure*}

\noindent\textbf{GenHowTo.}
GenHowTo is a large-scale dataset automatically constructed with 200K image triplets and corresponding textual descriptions.
Each triplet consists of temporally ordered video frames depicting (i) the initial state of an object, (ii) the action that modifies the state, and (iii) the resulting new state of the object.
These structured triplets, along with their descriptions, naturally align with interleaved text-image training for modeling step-by-step visual reasoning.
We utilize the provided action prompts and state prompts, along with their corresponding images, to construct a series of interleaved text-image samples for visual reasoning tasks, as shown in Figure~\ref{fig-data}.

\noindent\textbf{OpenStory++.}
OpenStory++ is a comprehensive dataset designed to emphasize narrative continuity around key instances, featuring instance-level visual segmentation.
It processes video content by extracting keyframes, evaluating them for aesthetic quality, and generating descriptive captions using BLIP2.
These captions are then refined by a Large Language Model (LLM) to ensure coherence and maintain narrative flow.
The resulting keyframe-caption pairs are used to build interleaved text-image sequences that support visual storytelling, as shown in Figure~\ref{fig-data}.

\begin{figure*}[t]
    \begin{center}
        \includegraphics[width=1.0\linewidth]{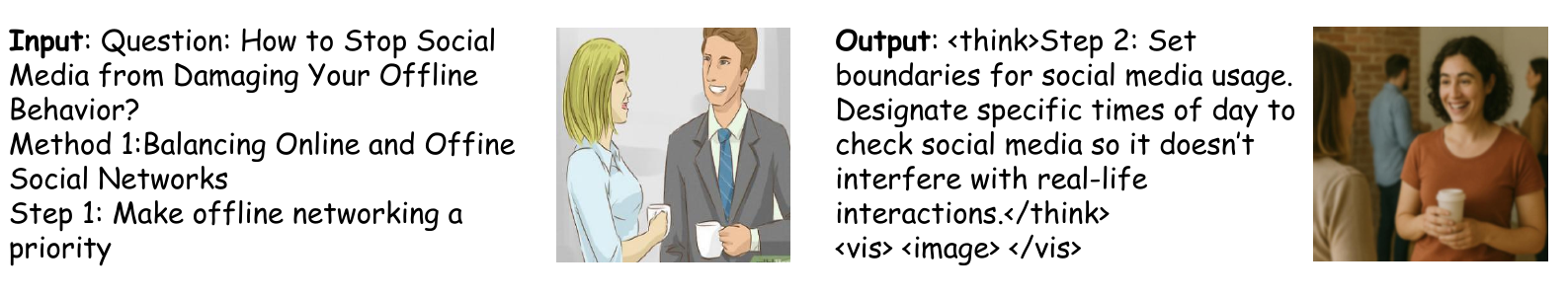}
    \end{center}
    \vspace{-2mm}
    \caption{Failure cases analysis about hallucinated visual content.}
    \label{fig-fcase1}
    \vspace{-2mm}
\end{figure*}

\section{Failure cases analysis}
While our method demonstrates strong performance on interleaved multimodal generation tasks, we observe several failure cases that highlight current limitations, as demonstrated in Figure~\ref{fig-fcase1}.
First, in some complex reasoning scenarios, the model may generate hallucinated visual content that does not faithfully correspond to the textual context, indicating limitations in cross-modal alignment.
Second, the visual elements produced by the model occasionally fail to match the style or appearance of the input image, resulting in stylistic inconsistency across modalities.
These issues suggest future work could benefit from improved reward modeling that better captures factual consistency, reasoning traceability, and structural fidelity in interleaved generation.

\section{Broader impacts}
The proposed work has the potential to greatly enhance the capability of unified models for multimodal interleaved generation.
Therefore it could be beneficial for applications such as multimodal reasoning and visual storytelling.
It could also be used to provide a better experience of multimodal interleaved interaction between human and AI system.
As for the potential negative impacts, there is a risk that the technology could be misused to generate and spread disinformation.

\end{document}